\documentclass[letterpaper, 10pt, conference]{ieeeconf}
\IEEEoverridecommandlockouts
\usepackage{cite}
\usepackage{amsmath,amssymb,amsfonts}
\usepackage{algorithmic}
\usepackage{graphicx}
\usepackage{textcomp}
\usepackage{xcolor}
\usepackage[labelfont=bf,compatibility=false]{caption}
\usepackage{subcaption}
\usepackage{booktabs}
\usepackage{hyperref} 
\usepackage{comment}
\usepackage{tabularx} 

\usepackage{siunitx}

\let\oldthebibliography\thebibliography
\let\endoldthebibliography\endthebibliography
\renewenvironment{thebibliography}[1]{%
  \oldthebibliography{#1}%
  \setlength{\itemsep}{0pt}%
  \setlength{\parskip}{0pt}%
}{\endoldthebibliography}

\begin{document}

\title{Towards Learning Boulder Excavation with Hydraulic Excavators 
}
\author{Jonas Gr\"utter$^1$, Lorenzo Terenzi$^1$, Pascal Egli$^2$ and Marco Hutter$^1$%
\thanks{$^1$The authors are with Robotic Systems Lab, ETH Zurich, 8092 Zurich, Switzerland.}%
\thanks{$^2$The author is with Gravis Robotics AG, 8050 Zurich, Switzerland.}%
\thanks{Corresponding author: Jonas Gruetter {\tt\small jgruette@ethz.ch}}%
\thanks{This work was supported by the NCCR Digital Fabrication (DFAB) and the Swiss National Science Foundation (SNF) through project No. 188596.}%
}

\maketitle

\vspace{-3mm} 


\begin{abstract}
Construction sites frequently require removing large rocks before excavation or grading can proceed. Human operators typically extract these boulders using only standard digging buckets, avoiding time-consuming tool changes to specialized grippers.
This task demands manipulating irregular objects with unknown geometries in harsh outdoor environments where dust, variable lighting, and occlusions hinder perception. The excavator must adapt to varying soil resistance—dragging along hard-packed surfaces or penetrating soft ground—while coordinating multiple hydraulic joints to secure rocks using a shovel.
Current autonomous excavation focuses on continuous media (soil, gravel) or uses specialized grippers with detailed geometric planning for discrete objects. These approaches either cannot handle large irregular rocks or require impractical tool changes that interrupt workflow.
We train a reinforcement learning policy in simulation using rigid-body dynamics and analytical soil models. The policy processes sparse LiDAR points (just 20 per rock) from vision-based segmentation and proprioceptive feedback to control standard excavator buckets. The learned agent discovers different strategies based on soil resistance: dragging along the surface in hard soil and penetrating directly in soft conditions.
Field tests on a 12-ton excavator achieved 70\% success across varied rocks (0.4--0.7m) and soil types, compared to 83\% for human operators. This demonstrates that standard construction equipment can learn complex manipulation despite sparse perception and challenging outdoor conditions.
\end{abstract}


\vspace{-2mm} 

\section{Introduction}

Autonomous boulder removal remains a critical gap in construction robotics. During site preparation, operators extract rocks using standard digging buckets, relying on visual and tactile feedback to adapt to varying rock geometries and soil conditions.

This capability is essential for practical automation. While grippers offer better control, tool changes interrupt workflow and are impractical when the bucket is needed for subsequent excavation. Manipulating irregular objects with standard buckets would enable seamless task transitions.

Boulder extraction requires coordinating multiple joints to align, scoop, and secure irregularly shaped rocks while adapting to soil resistance—from hard surfaces where rocks slide to soft conditions requiring different penetration depths. Outdoor environments compound these challenges through dust, variable lighting, long-range sensing requirements (\SIrange{7}{10}{\meter}), and arm-induced occlusions.

Recent work demonstrates autonomous excavation of continuous media \cite{egli2022soil, egli2024reinforcement} and precise assembly with discrete objects using grippers \cite{johns2023framework, 9392354}. Learning-based methods adapt trajectories using hydraulic feedback \cite{egli2022soil}, while classical approaches use specialized grippers to calculate stable grasp points on detailed 3D models of objects \cite{johnsAutonomousDryStone2020, wermelingerGraspingObjectReorientation2021}.

\begin{figure}[t]
\vspace{-2mm} 
    \centering
    \includegraphics[width=\columnwidth]{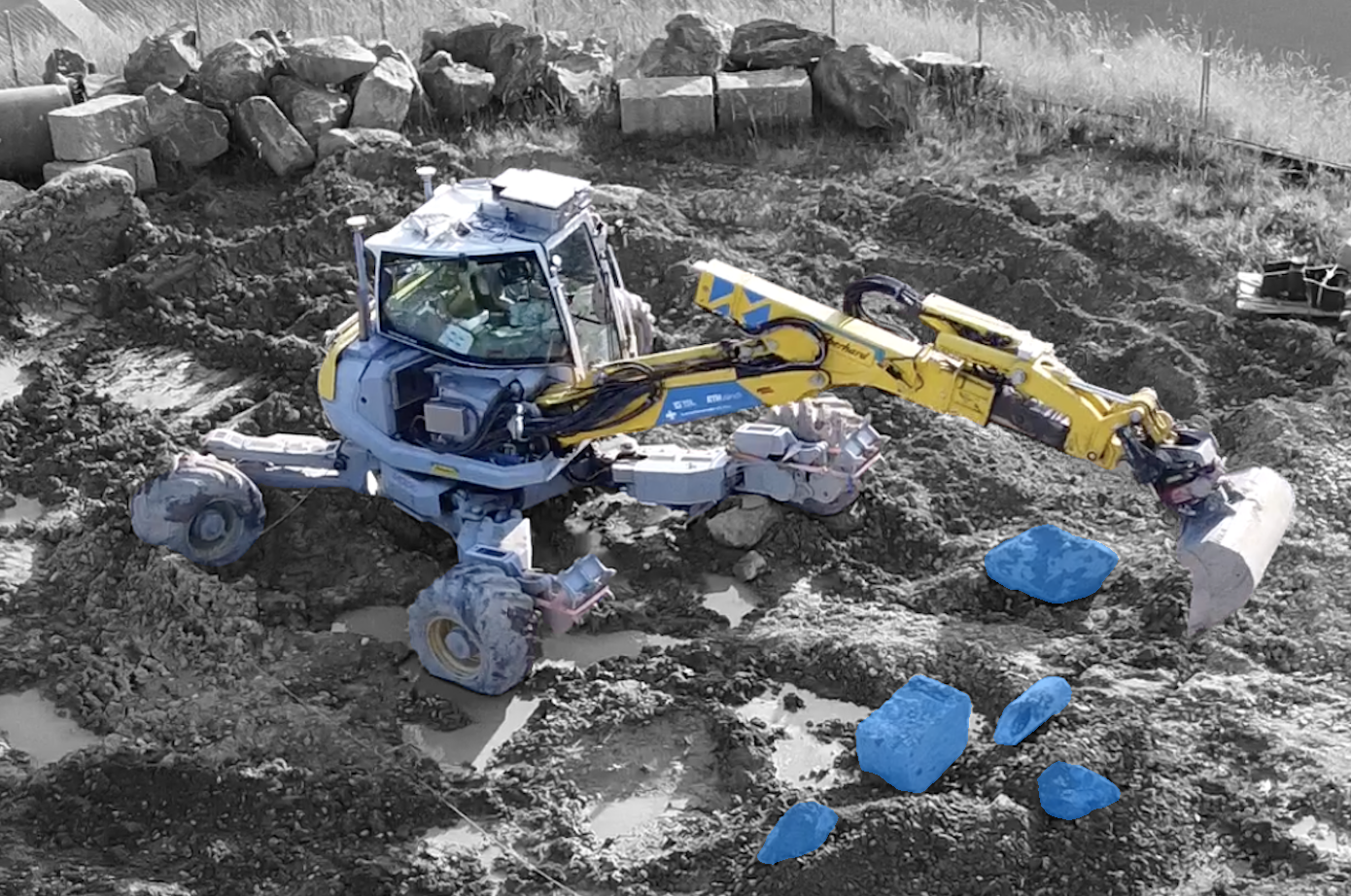}
    \caption{Example excavation scene with rocks of varying shapes, sizes, and surface conditions (target rocks highlighted in blue). Reliable detection and removal of such objects is critical for safe and efficient downstream tasks like terrain leveling or continued excavation.}
    \label{fig:picturemain}
\vspace{-3mm} 
\end{figure}

These approaches fail for boulder extraction with standard tools. Classical methods need accurate 3D models and specialized end-effectors. Continuous media controllers lack reactive behaviors for individual object manipulation. No existing method handles the combination of irregular objects, standard buckets, and harsh outdoor conditions.

We present a reinforcement learning approach enabling autonomous boulder excavation using standard buckets and sparse perception—just 20 LiDAR points per rock.

Policies trained in simulation with rigid-body dynamics and analytical soil models learn adaptive strategies from sparse point clouds and proprioceptive feedback. Different behaviors emerge without explicit soil estimation: surface dragging in hard soil versus direct penetration in soft conditions.

Field deployment on a 12-ton excavator achieved 70\% success across varied rocks (0.4--0.7m) and soil types, compared to 83\% for human operators, demonstrating complex manipulation without specialized tools.

Our contributions are:
\begin{enumerate}
\item A reinforcement learning framework combining rigid-body simulation with analytical soil models to discover adaptive boulder pickup strategies using standard buckets.

\item A sparse perception pipeline using SAM2 segmentation with LiDAR, demonstrating boulder excavation with only 20 points per rock.

\item Field validation achieving 70\% success rate compared to 83\% for human operators across diverse conditions.
\end{enumerate}
\vspace{-1mm} 

\section{Related Work}
The convergence of advanced perception systems and learning-based control has opened new possibilities for autonomous manipulation across scales. While dexterous manipulation has seen rapid progress through deep learning, transferring these advances to heavy machinery presents unique challenges in perception, control, and safety. This work bridges these domains, presenting the first demonstration of robust boulder manipulation on construction equipment using only a standard digging bucket, guided by reinforcement learning and sparse 3D perception.

\subsection{Perception-based Learning for Manipulation}
\label{sec:perception}
Learning-based manipulation has progressed from RGB-based policies \cite{levine2018learning, kalashnikov2018qt} to 3D representations better suited for real-world robotics. Early approaches using pre-trained vision encoders like CLIP \cite{shridhar2023perceiver, intelligence2025pi05} often fail in outdoor settings due to depth ambiguity and poor generalization. For construction machinery operating at long range, point clouds provide more reliable spatial information.

LiDAR is ideal for construction sites due to its robustness to lighting conditions and its ability to capture large-scale 3D structures. Macaro et al. \cite{macaro} employ dual LiDARs on an excavator to enable rock segmentation via geometric clustering; however, performance degrades on uneven terrain where boundaries become ambiguous or individual rocks are over-segmented.

Point cloud processing has advanced from PointNet's \cite{qi2017pointnet} unordered input handling to hierarchical models \cite{qi2017pointnet2} and recent RL-tailored designs. PointPatchRL \cite{gyenes2024pointpatchrl} introduces patch-based transformers, while ANYmal parkour \cite{hoeller2024anymal} uses 4D convolutions on LiDAR data, favoring this modality over depth cameras for its superior range—a key advantage in our setup.

Many complex manipulation policies rely on demonstration data. Imitation learning approaches such as RVT-2 \cite{goyal2024rvt2} and Act3D \cite{gervet2023act3d} employ large 3D encoders trained on expert rollouts, but collecting such data is often impractical for heavy machinery. Distillation-based methods~\cite{chen2023visual, ze2024h} also depend on high-quality visual inputs, which prove unreliable in outdoor environments due to dust, occlusion, and variable lighting conditions.

\subsection{Autonomous Heavy Machinery and Construction~Robotics}
\label{sec:heavymachinery}
Construction automation has traditionally focused on continuous media tasks such as trenching \cite{cannonExtendedEarthmovingAutonomous1999}, grading \cite{judRoboticEmbankment2021}, and loading \cite{stentzRoboticExcavatorAutonomous1998, zhangAutonomousExcavatorSystem2021a}. More recently, efforts have extended to manipulating discrete, irregular objects. Classical approaches to such tasks, including autonomous dry stone wall assembly~\cite{johnsAutonomousDryStone2020, wermelingerGraspingObjectReorientation2021}, rely on gripper-based grasp planning with force-closure guarantees \cite{9392354} and detailed perception pipelines for precise placement. While effective in structured settings, these methods require accurate 3D models and hand-crafted grasp and placement strategies tailored to specific object geometries, as well as specialized end~effectors.

In contrast, tasks like site preparation demand fast, robust boulder excavation using standard tools. Here, strategies must adapt to uncertain rock shapes and varying soil resistance, making reactive behavior more critical than geometric precision.

Learning-based methods address this challenge by enabling adaptive behavior to emerge from interaction, eliminating the need for explicit modeling or manual heuristics. For example, soil-adaptive excavation \cite{egli2022soil} demonstrated that reinforcement learning (RL) policies can adjust trajectories based on hydraulic pressure feedback, while autonomous bucket filling \cite{egli2024reinforcement} showed that proprioceptive signals support material-agnostic adaptation.

Building on these insights and the perception advances discussed in Section~\ref{sec:perception}, the application of RL to point-cloud inputs proves crucial. For instance, recent work \cite{perceptiveexcavation} trains RL policies using compact geometric point-cloud embeddings and lightweight MLPs for sequential excavation of cluttered rigid objects, thereby reducing both training time and policy complexity. Unlike Lu et al.~\cite{perceptiveexcavation}, who focus on sequential excavation of small rigid objects (1-7cm) using 7-DOF robotic arms with rich geometric representations, our approach tackles boulder excavation—manipulating massive irregular rocks (0.4-0.7m diameter) on industrial-scale 12-ton excavators using extremely sparse perception (only 20 LiDAR points) and analytical soil interaction models for field deployment on construction equipment.

\section{Method}
\begin{figure}[b]
\vspace{-2mm}
    \centering
    \includegraphics[width=\columnwidth]{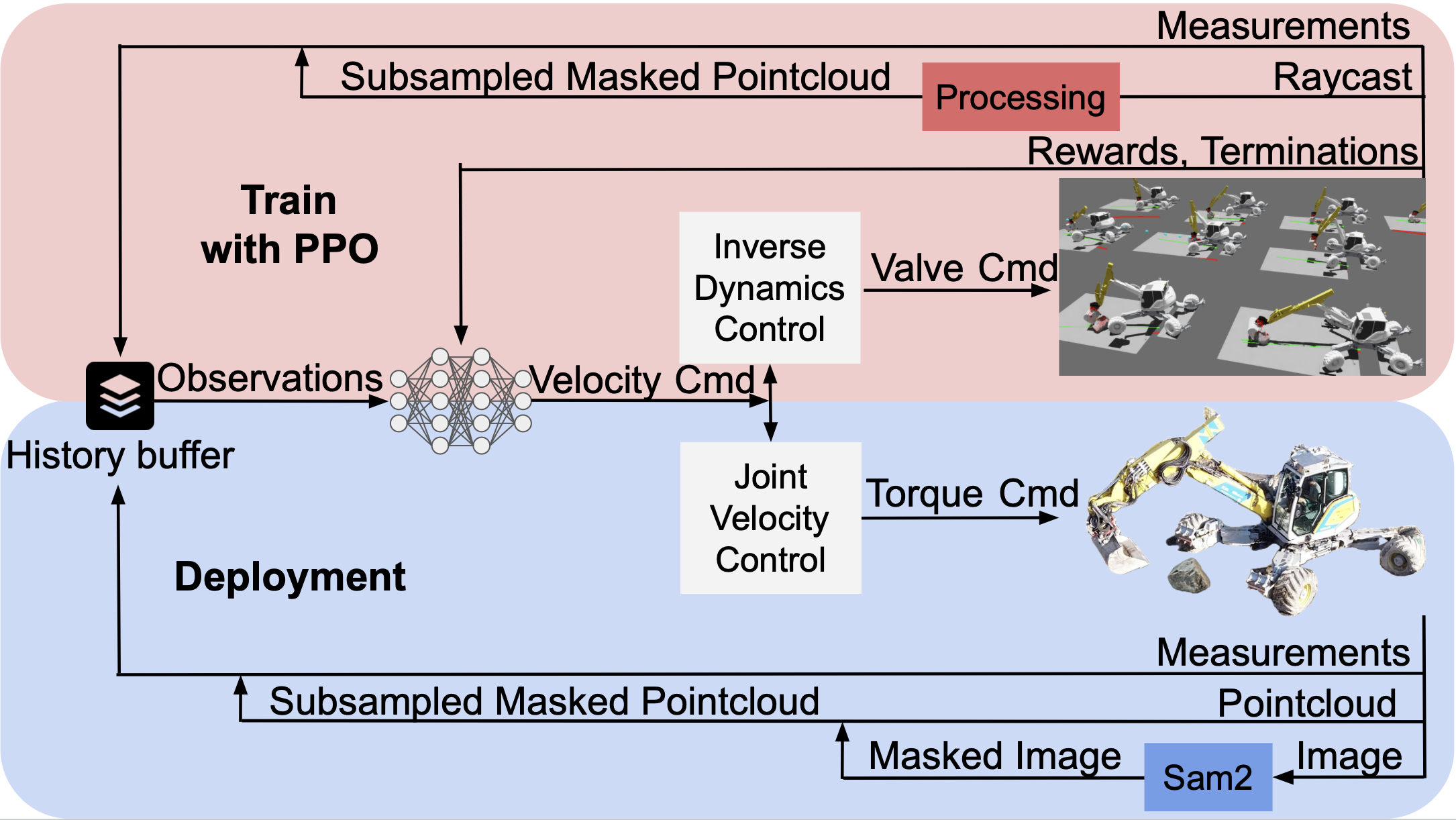}
    \caption{System architecture overview, illustrating the training loop (top) and the deployment pipeline (bottom). The perception pipeline processes RGB images through SAM2 segmentation combined with sparse LiDAR points to enable autonomous boulder extraction using standard excavator buckets.}
    \label{fig:pipeline}
\vspace{-3mm}
\end{figure}

We train the policy in simulation using a soil resistance model and sparse point cloud observations, then deploy it on a 12-ton hydraulic excavator. The full system—including perception, simulation, and control—is summarized in Fig.~\ref{fig:pipeline}.

\subsection{Simulation}
\label{sec:simulation}

\begin{figure}[htbp]
\vspace{-2mm}
    \centering
    \includegraphics[width=\columnwidth]{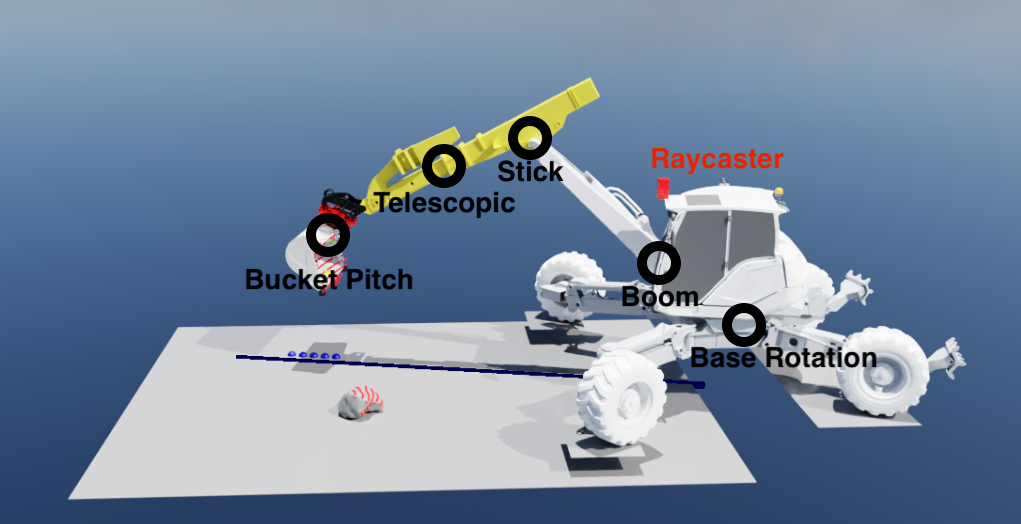}
    \caption{Simulation setup. Red points indicate LiDAR raycast hits, and the dark blue line represents the soil surface model aligned with the excavator arm.}
    \label{fig:setup}
\vspace{-3mm}
\end{figure}

The training environment for our RL agent was developed within NVIDIA Isaac Lab \cite{Mittal_2023}.
Isaac Lab is used to simulate the rigid-body dynamics of the excavator and its interaction with rocks and terrain, excluding soil forces which are modeled separately.
\\
\textit{Excavator Model:} The excavator model was simplified to include only the key actuated joints—base rotation, boom, stick, telescopic extension, and bucket pitch—to enable efficient simulation (Fig.~\ref{fig:setup}). Remaining joints were fixed, and hydraulic dynamics were omitted. Hydraulic dynamics are handled by low-level PID controllers on the real hardware that track velocity commands, as detailed in Section~III-B. The excavator stands on three static support bases to emulate a fixed, stable operating position with a friction coefficient of 0.8 \cite{egli2024reinforcement}.
\\
\textit{Rocks:} 
A dataset of 50 real rock geometries was generated by 3D scanning physical rocks, capturing natural variation in shape and surface features. Each rock was assigned a nominal density of \SI{2500}{\kilogram\per\cubic\meter}, with mass randomized by \SI{+-10}{\percent} per episode. Dimensions were varied along the principal axes (\textit{x}, \textit{y}, \textit{z}) within the ranges \SIrange{0.3}{1.0}{\meter}, \SIrange{0.1}{0.5}{\meter}, and \SIrange{0.2}{0.7}{\meter}, respectively. Friction coefficients were sampled uniformly from \num{0.35} to \num{0.6}, based on reported values for dry rock–soil interfaces. Rock instances were randomized across parallel environments to promote generalization.
\\
\textit{Environment Setup:} Rocks are instantiated on a flat, horizontal support surface. Collision properties are defined such that the excavator's bucket can collide with the rocks, and the rocks can collide with both the bucket and the support surface. However, the excavator body itself does not collide with this support platform. This configuration creates a designated manipulation surface, akin to a testbed table, where rocks are presented for the excavation task. 
\\
\textit{Soil Model:} The bucket–soil interaction is simulated using a quasi-static 2D excavation force model, implemented following~\cite{egli2022soil}, and based on the classical Fundamental Equation of Earth-Moving \cite{fee}. The model estimates resistive forces using key geotechnical parameters—soil cohesion, internal friction angle, and soil-tool adhesion—computed in the plane of excavation. The 2D simplification enables computationally efficient training across thousands of parallel environments required for RL, while parameter randomization bridges the fidelity gap by exposing the policy to diverse soil conditions. These parameters are randomized during training to encourage generalization across soil types. The soil is modeled as a planar layer beneath the rock, co-located with the support platform. Interaction forces are computed in a vertical slice aligned with the current arm configuration, and applied to the shovel in Isaac Lab as state-dependent external forces opposing the digging trajectory. The "soft" and "hard" soil conditions referenced in evaluation correspond to the specific parameter ranges shown in Table~\ref{tab:soil_parameters}.

\begin{table}[htbp]
\vspace{-2mm}
\centering  
\caption{Geotechnical parameters for soft and hard soil conditions used in simulation and evaluation}
\label{tab:soil_parameters}
\footnotesize
\begin{tabular}{@{}lcc@{}}
\toprule
\textbf{Parameter} & \textbf{Soft Soil} & \textbf{Hard Soil} \\
\midrule
Cohesion, $c$ [Pa] & 0 & 105,000 \\
Internal friction angle, $\phi$ [°] & 31.5 & 32.0 \\
Unit weight, $\gamma$ [N/m³] & 19,500 & 21,000 \\
Soil-metal friction angle, $\delta$ [°] & 23.0 & 23.0 \\
Cavity expansion factor, $CP$ & 1 & 300 \\
Adhesion factor, $c_a/c$ & 0 & 0.5 \\
\bottomrule
\end{tabular}
\vspace{-2mm}
\end{table}

\textit{Simulated Perception}: LiDAR-like point clouds are generated via Isaac Lab's dynamic raycasting, using a virtual sensor mounted atop the cabin to mimic the real hardware's position and field of view. Rays intersect with both rock and bucket meshes, simulating occlusions. The simulator tags each hit with its source mesh, enabling the extraction of points belonging to the target rock, which are then used as policy input.

\subsection{Baseline}
The baseline is the continuous soil excavation policy from Egli et al.~\cite{egli2022soil, egli2024reinforcement}, which outputs joint velocity commands using identical soil resistance modeling. However, it was trained for continuous media excavation with observations focused on soil height fields and rewards targeting soil volume collection. 

\subsection{Reinforcement Learning Formulation}

\label{Sec Reinforcement Learning Formulation}

The agent learns a policy $\pi(a_t|s_t)$ that maps observations $s_t$ to joint velocity commands $a_t$ to acquire a target rock. This involves: (1) aligning the bucket with the rock, (2) executing a scooping motion to capture it, and (3) lifting and curling to secure it. The actuated joints include base turn, boom, stick, telescopic extension, and bucket pitch, denoted $q = [q_{\text{turn}}, q_{\text{boom}}, q_{\text{stick}}, q_{\text{tele}}, q_{\text{pitch}}]^T$.

\subsubsection[Observation Space]{Observation Space ($s_t$)}
The observation vector concatenates the following components:
\begin{itemize}
\setlength{\itemsep}{0pt}
\item Proprioceptive state: joint positions, velocities, and torques as in~\cite{egli2022soil}
\item Turn joint position and its velocity history of length $L$
\item Bucket pose and velocity in the rotating cabin frame
\item Downsampled point cloud of $N=20$ 3D points representing the target rock in the rotating base frame
\item Previous action $a_{t-1}$, with turn joint actions stored in a history buffer of length $L$
\end{itemize}
All inputs are normalized to $[-1, 1]$.

\subsubsection[Action Space]{Action Space ($a_t$)}
The policy outputs joint velocity commands:
\begin{equation}
    a_t = [\dot{q}_{\text{turn}}, \dot{q}_{\text{boom}}, \dot{q}_{\text{stick}}, \dot{q}_{\text{tele}}, \dot{q}_{\text{pitch}}]^T
\end{equation}
Turn commands include a deadband threshold $\epsilon_{\text{turn}}$:
\begin{equation}
\label{eq:deadzone}
    \dot{q}_{\text{turn}} =
    \begin{cases}
        0 & \text{if } |\dot{q}_{\text{turn}}| < \epsilon_{\text{turn}} \\
        \dot{q}_{\text{turn}} & \text{otherwise}
    \end{cases}
\end{equation}
To emulate real-world delays, a control delay $\tau_{\text{delay}} \in [0, 1.2]$ \si{\second} is randomly sampled at the start of each episode and applied throughout, enabling online delay compensation during training ~\cite{spinelli2024reinforcement}. The turn joint is particularly affected by delays due to its different mechanics compared to traditional hydraulic cylinders and strong dependence on excavator inertia from varying arm configurations. The observation history length $L$ is chosen such that $L \cdot \Delta t_{\text{control}} \geq \max(\tau_{\text{delay}})$, ensuring sufficient context for compensation.

\subsubsection{Rewards}
The reward function combines dense shaping and sparse terminal signals to guide efficient, safe, and task-completing behaviors, as summarized in Table~\ref{tab:rewards}. Rewards R1--R6 encourage proper alignment, successful scooping, curling beyond $\theta_{\text{target}} = \SI{0.5}{\radian}$, lifting to a target height $h_{\text{desired}} = \SI{0.5}{\meter}$ above the soil. Penalties P1--P2 discourage abrupt actions and excessive bucket velocities above $v_{\text{max}} = \SI{0.6}{\meter\per\second}
$, while P3--P4 penalize unsafe or unnecessary turning, particularly during soil interaction. R7 and P6 serve as terminal rewards or penalties based on task outcome. 

\begin{table*}[t]
\centering
\caption{Reward and penalty terms used during training. \textit{Notation.}
We denote positions and velocities using $\mathbf{p}_{\text{object}}^{\text{frame}}$ and $\mathbf{v}_{\text{object}}^{\text{frame}}$, respectively, where the superscript indicates the reference frame (e.g., base, rotbase, world). For instance, $\mathbf{v}_{\text{bucket}}^{\text{rotbase}}$ is the linear velocity of the bucket in the rotating base frame. The bucket's curl angle is $\theta_{\text{curl}}$, with $\theta_{\text{target}}$ the desired value for securing a rock. The digging penalty P5 depends on the current bucket depth $\text{d}_{\text{bucket}}$, and thresholds $\text{d}_{\text{soft}}$ and $\text{d}_{\text{hard}}$, along with lateral alignment $y_{\text{bucket}}$ with bounds $y_{\text{min}}$ and $y_{\text{max}}$.
}
\label{tab:rewards}
\renewcommand{\arraystretch}{1.2}
\small
\begin{tabular}{@{}llp{6cm}c@{}}

\toprule
\textbf{ID} & \textbf{Description} & \textbf{Expression} & \textbf{Weight} \\
\midrule
\multicolumn{4}{c}{\textbf{Rewards}} \\
\midrule
R1 & Alignment with boulder (lateral) & $\exp\left(-|\mathbf{p}_{\text{boulder}, y}^{\text{rotbase}}|^2\right)$ & $0.005$ \\[0.3ex]

R2 & Bucket proximity to boulder & $\mathbf{1}_{\|\mathbf{p}_{\text{boulder}, y}^{\text{rotbase}} - \mathbf{p}_{\text{bucket}, y}^{\text{rotbase}}\|^2 < 1.5}$ & $0.01$ \\[0.3ex]

R3 & Bucket beneath boulder & $\mathbf{1}_{\text{R2} \neq 0} \cdot \mathbf{1}_{\substack{\mathbf{p}_{\text{bottom plate}, z}^{\text{base}} < \\ \mathbf{p}_{\text{boulder}, z}^{\text{base}}}}$ & $0.01$ \\[0.4ex]

R4 & Rock inside shovel & $\mathbf{1}_{\text{boulder in shovel}}$ & $0.075$ \\[0.3ex]

R5 & Rock secured via curl & $\mathbf{1}_{\text{in shovel}} \cdot \mathbf{1}_{\text{curled enough}}$ & $0.05$ \\[0.3ex]

R6 & Lifted to target height & $\mathbf{1}_{\Delta z > 0} \cdot \mathbf{1}_{\text{in shovel}} \cdot \exp\left(-\left\|\mathbf{p}_{\text{bucket}, z}^{\text{base}} - h_{\text{desired}}\right\|^2\right)$ & $0.05$ \\[0.4ex]

R7 & Task success (terminal) & $\mathbf{1}_{\text{success}}$ & $20$ \\[0.3ex]

\midrule
\multicolumn{4}{c}{\textbf{Penalties}} \\
\midrule

P1 & Action smoothness & $\left\|\mathbf{a}_t - \mathbf{a}_{t-1}\right\|^2$ & $-0.005$ \\[0.3ex]

P2 & Excessive bucket velocity & $\max\left(0, \left\|\mathbf{v}_{\text{bucket}}^{\text{rotbase}}\right\| - v_{\text{max}}\right) \cdot 10^{\substack{\max(0, \|\mathbf{v}_{\text{bucket}}^{\text{rotbase}}\| \\ - v_{\text{max}})}}$ & $-0.1$ \\[0.4ex]

P3 & Unnecessary turn & $\mathbf{1}_{|\dot{q}_{\text{turn}}| \geq \epsilon_{\text{turn}}}$ & $-0.005$ \\[0.3ex]

P4 & Turning while digging & $\mathbf{1}_{\text{in soil}} \cdot \mathbf{1}_{|\dot{q}_{\text{turn}}| \geq \epsilon_{\text{turn}}}$ & $-0.025$ \\[0.3ex]

P5 & Misaligned shallow scooping & $\frac{d_{\text{bucket}} - d_{\text{soft}}}{d_{\text{hard}} - d_{\text{soft}}} \cdot \frac{|y_{\text{bucket}}| - y_{\text{min}}}{y_{\text{max}} - y_{\text{min}}}$ & $-0.0125$ \\[0.4ex]

P6 & Task failure (terminal) & $\mathbf{1}_{\text{failed}}$ & $-0.5$ \\[0.3ex]
\bottomrule
\end{tabular}
\vspace{-2mm}
\end{table*}

\subsubsection{Termination Conditions}

Episodes terminate on task success, safety violations, or timeout, enabling clear supervision and correct advantage estimation with GAE~\cite{schulman2018highdimensionalcontinuouscontrolusing}. Terminations fall into three categories: T1 (timeout after $\SI{29}{\second}$), T2--T6 (safety and physical constraints), and T7 (success: the boulder is lifted above $h_{\text{desired}}$ and curled beyond $\theta_{\text{target}}$s). Safety thresholds include a maximum base speed of $v_{\text{max, base}} = \SI{0.1}{\meter\per\second}$, bucket velocity limit $v_{\text{max, term}} = \SI{1.2}{\meter\per\second}$, and angle-of-attack threshold $\alpha_{\text{threshold}} = \SI{0}{\radian}$. The boulder is considered dropped if it falls below $h_{\text{min}}$, defined by platform height. See Table~\ref{tab:terminations} for full criteria.
\begin{table}[htbp]
\centering
\caption{Termination conditions used during training and evaluation. \textit{Notation.} $\alpha$ denotes the angle between the bucket's bottom plate orientation and its instantaneous velocity vector in the $xz$ plane, capturing the alignment between the cutting edge and its direction of motion. Time limit is $\SI{29}{\second}$.}
\label{tab:terminations}
\renewcommand{\arraystretch}{1.1} 
\footnotesize 
\begin{tabularx}{\columnwidth}{@{}lX@{}}
\toprule
\textbf{Condition} & \textbf{Criterion} \\
\midrule
T1: Episode timeout & 
$\text{time} \geq \text{time\_limit}$ \\

T2: Excessive base velocity (-) & 
$\|\mathbf{v}_{\text{base}}^{\text{world}}\| > v_{\text{max, base}}$ \\

T3: Joint velocity limit exceeded (-) & 
$\max_i |\dot{q}_i| > \dot{q}_{\text{max}}$ \\

T4: Excessive bucket speed (-) & 
$\|\mathbf{v}_{\text{bucket}}^{\text{rotbase}}\| > v_{\text{max, term}}$ \\

T5: Boulder dropped (-) & 
$\mathbf{p}_{\text{boulder}, z}^{\text{base}} < h_{\text{min}}$ \\

T6: Invalid angle of attack (-) & 
$\alpha_{\text{bucket}} > \alpha_{\text{threshold}}$ \\

T7: Task success (+) & 
$\text{boulder in shovel} \land (\theta_{\text{curl}} > \theta_{\text{target}}) \land (\mathbf{p}_{\text{boulder}, z}^{\text{base}} > h_{\text{desired}})$ \\
\bottomrule
\end{tabularx}
\vspace{-1mm}
\end{table}

\subsubsection{Curriculum and Initialization}

To accelerate training and improve generalization, we employ a five-stage curriculum that progressively increases task complexity. Each episode begins from a pre-generated, collision-free configuration sampled from a cached dataset of valid initial states, ensuring safe and consistent resets.

The curriculum is structured to guide the agent through increasingly difficult behaviors—starting with lifting, then scooping, alignment, and ultimately generalizing across diverse rock and soil configurations.

\vspace{-1mm}
\begin{itemize}
\setlength{\itemsep}{0pt}
    \item \textbf{Level 0:} Rock is placed directly in front of the bucket (within a $2.5 \times 0.6 \times 0.4$~m region); soil is soft; angle-of-attack termination (T6) is disabled to focus on learning lifting without caring how the soil is penetrated.
    
    \item \textbf{Level 1:} T6 is activated, enforcing correct bucket orientation during scooping.

    \item \textbf{Level 2:} Soil properties (cohesion, adhesion factor, internal friction angle, soil-tool friction angle, weight unit per volume, cutting resistance) are randomized, exposing the agent to varying excavation dynamics.

    \item \textbf{Level 3:} Rock and bucket initial positions are sampled from a wider $4.5 \times 3 \times 0.4$~m volume, requiring learning alignment strategies.

    \item \textbf{Level 4:} P5 penalty is enabled to discourage misaligned digging; rock longitudinal position is further randomized to train depth-aware scooping.
\end{itemize}
\vspace{-1mm}

Transitions between levels are triggered automatically based on training performance: the agent advances to the next level once the success rate on the current level exceeds 80\% over a rolling window of 1000 episodes. This staged approach enables efficient skill acquisition and supports robust sim-to-real transfer.

\subsubsection{Training}
The policy is trained using Proximal Policy Optimization (PPO)~\cite{schulman2017proximalpolicyoptimizationalgorithms} with Generalized Advantage Estimation (GAE)~\cite{schulman2018highdimensionalcontinuouscontrolusing}. We use an actor-critic architecture with two hidden layers of 256 units, leaky ReLU activations, and an initial noise standard deviation of 0.4. Optimization uses a fixed learning rate of $10^{-4}$, clip ratio 0.2, value loss coefficient 0.5, entropy coefficient 0.005, and 5 epochs per update over 4 minibatches. The agent is trained for 9000 iterations (24 steps per environment per update), requiring approximately 8 years of experience in simulation. Full training code is based on RSL-RL~\cite{rudinLearningWalkMinutes2022}.

\subsection{Hardware Platform and Perception Pipeline}

We deploy our policy on a Menzi Muck M545, a 12-ton hydraulic walking excavator~\cite{Jud_2021}. The learned joint velocity commands from simulation are executed on the real system using PID and feedforward valve-flow controllers, which generate solenoid currents for the hydraulic valves. Autonomous control is enabled by retrofitting electrically actuated pilot-stage valves. Arm joint states are measured via IMUs, leg joints with magnetostrictive sensors, and joint torques are estimated from hydraulic cylinder pressures. An onboard IMU provides global orientation.

Perception is performed by a cabin-mounted Ouster OS1 Rev 6 LiDAR and a ZED 2i stereo camera. We employ SAM2~\cite{ravi2024sam2}, a foundation model with video tracking capabilities, to segment the target rock. A custom ROS~2 wrapper enables real-time tracking within a moving window~\cite{myproject}. Segmentation is manually initialized by the user, who selects a small number of pixels (5) on the target rock in the camera image. The resulting mask is back-projected onto the LiDAR point cloud to extract a 3D surface representation. Inference runs onboard on an NVIDIA RTX 4080 GPU with a latency of $\simeq 95$~ms. Given our control loop operates at 6 Hz, this allows one inference per policy step.

\begin{figure}[htbp]
\vspace{-2mm}
  \centering
  \begin{subfigure}[t]{0.47\columnwidth} 
    \centering
    \includegraphics[width=\linewidth]{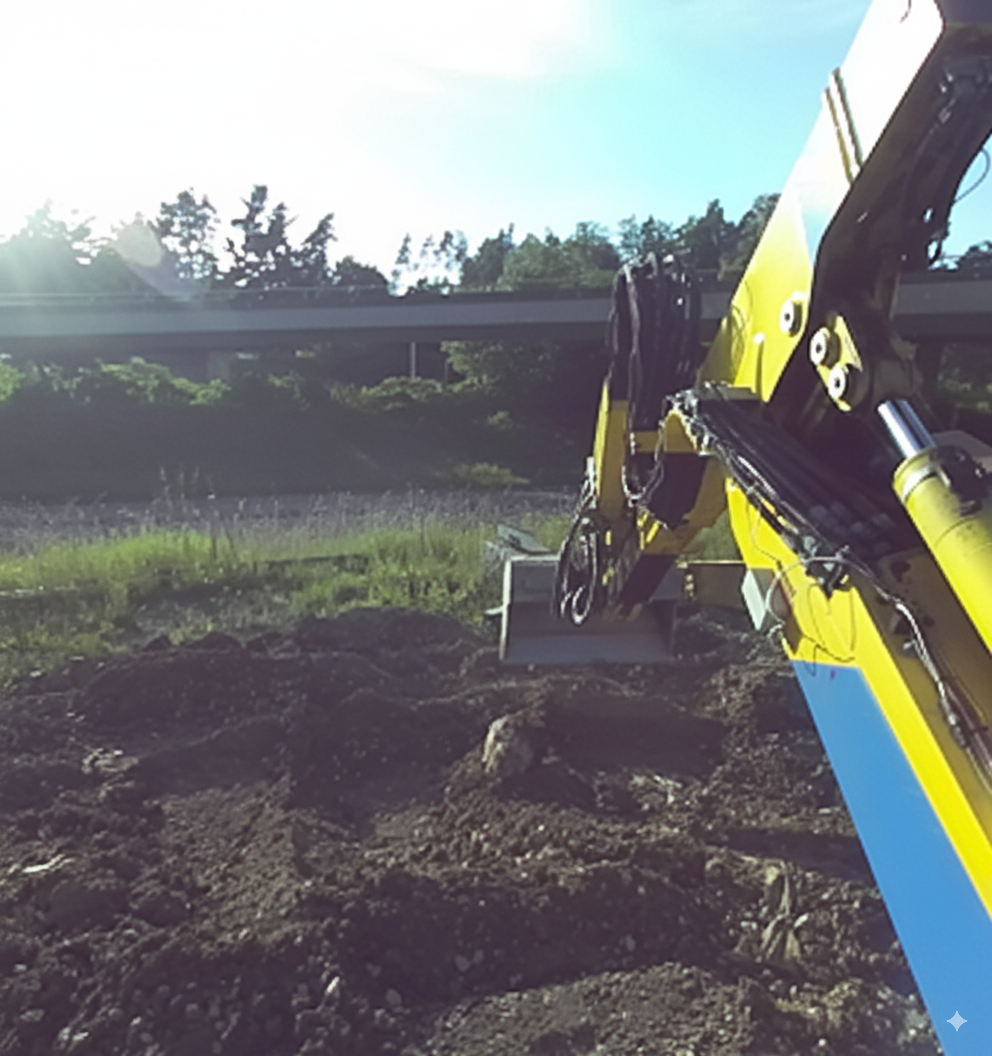}
    \caption{Raw image}
    \label{fig:side_a}
  \end{subfigure}
  \hfill
  \begin{subfigure}[t]{0.47\columnwidth} 
    \centering
    \includegraphics[width=\linewidth]{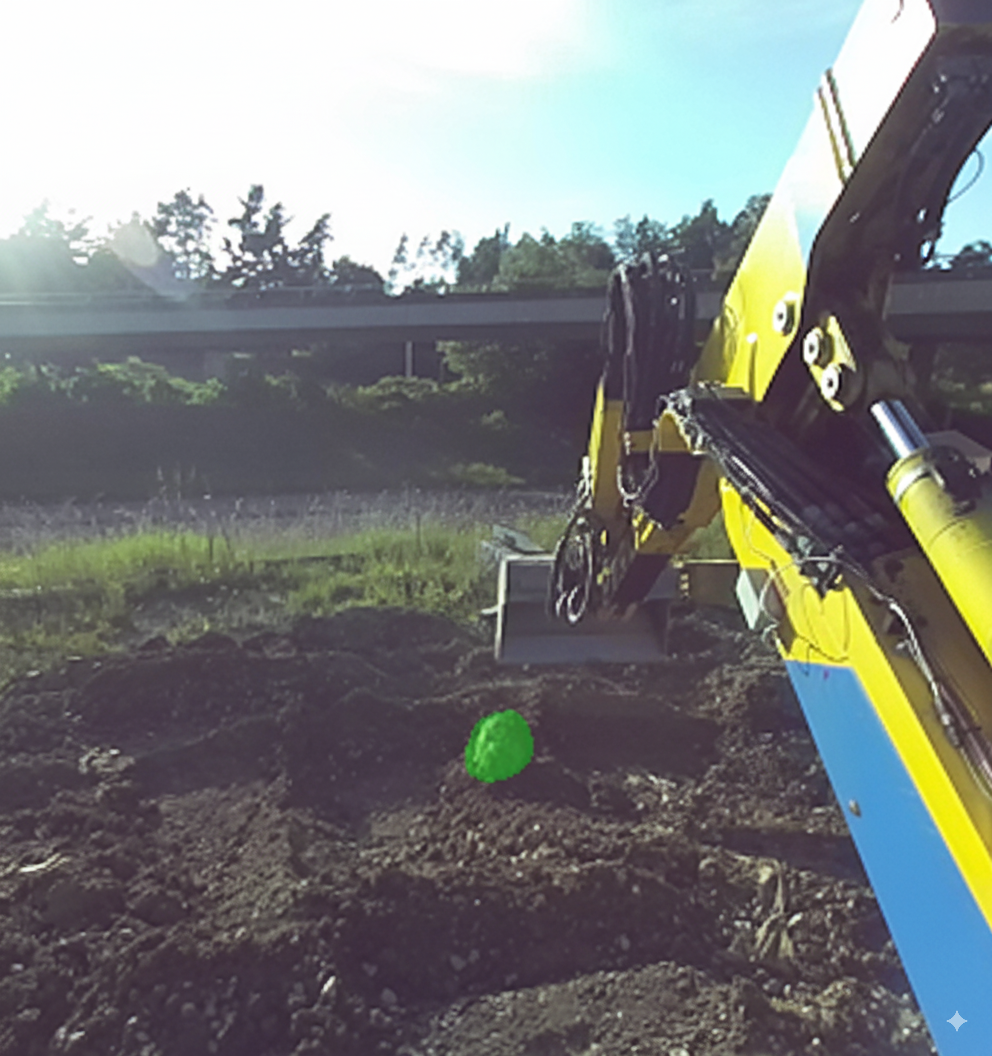}
    \caption{Masked image after user initialization}
    \label{fig:side_b}
  \end{subfigure}
  \caption{SAM2 segmentation in a difficult outdoor setting: low light, occlusion, and cluttered terrain.}
  \label{fig:sam2example}
\vspace{-3mm}
\end{figure}

SAM2 is well-suited for the harsh outdoor conditions encountered (mud, shadows, diverse rock textures) due to its strong generalization capabilities. Its robustness across lighting and occlusion scenarios enables zero-shot segmentation in challenging real-world scenes (Fig.~\ref{fig:sam2example}). The full deployment pipeline is summarized in Fig.~\ref{fig:pipeline}.

 \section{Results}

  We evaluate our approach through extensive simulation experiments and field deployment on a 12-ton Menzi
  Muck M545 excavator. Results demonstrate effective boulder extraction across varied conditions, with
  simulation comparisons highlighting the necessity of specialized learning for this task.

  \subsection{Simulation Experiments}
  \label{sec:simulation_results}

Test conditions encompassed 50 scanned rock geometries divided into small (0.3-0.5m) and large (0.8-1.0m) categories, with soil parameters ranging from soft to hard. Our method demonstrated consistently high performance (88-97\%) across all conditions, while the baseline approach succeeded only with small rocks in soft soil (86\%), where standard digging trajectories naturally capture rocks along with surrounding soil. The baseline's performance degraded significantly with larger rocks even in soft soil (44\%) and failed entirely in harder soil conditions.

This performance gap reveals the fundamental complexity of boulder manipulation. In hard soil, effective scooping becomes impossible as rocks slide along the surface rather than entering the bucket when penetration depth is insufficient. These challenging conditions demand highly reactive behaviors—the ability to dynamically adapt trajectory and depth based on real-time rock pose feedback—capabilities that traditional continuous media controllers inherently lack. Large rocks present additional challenges: they require precise reactive maneuvers to fit properly within the bucket, which explains our method's slightly reduced but still robust performance (88\%) in the most demanding scenario.

To evaluate generalization, we tested on 25 previously unseen rock geometries (0.8-1.0m) in hard soil conditions. The system achieved 87\% success compared to 88\% on training rocks, demonstrating strong robustness to novel shapes and confirming that our approach learns generalizable manipulation strategies rather than memorizing specific geometries.

  \subsection{Hardware Deployment}
  \label{sec:hardware_results}

  We conducted 40 hardware trials across four conditions: two rock sizes (approximately 0.4m and 0.7m
  diameter) and two soil types (soft and hard). An operator with approximately 300 hours of general
  excavator experience performed the same tasks for comparison. We compare against the continuous excavation baseline from Egli et al.~\cite{egli2022soil, egli2024reinforcement}, which lacks boulder-specific manipulation capabilities. Tables~\ref{tab:simulation_results} and~\ref{tab:hardware_results} summarize
  the results.



\begin{table}[htbp]
\centering
\caption{Simulation Results}
\label{tab:simulation_results}
\small
\begin{tabular}{@{}lcc@{}}
\toprule
\textbf{Condition} & \begin{tabular}{c} \textbf{Boulder RL} \\ \textbf{(Ours)} \end{tabular} & \begin{tabular}{c} \textbf{Digging RL} \\ \textbf{Baseline} \end{tabular} \\
\midrule
Small/Soft  & 96\% & 86\% \\
Big/Soft    & 97\% & 44\% \\
Small/Hard  & 94\% & 4\%  \\
Big/Hard    & 88\% & 0\%  \\
\midrule
\textbf{Average} & \textbf{94\%} & \textbf{33\%} \\
\bottomrule
\end{tabular}
\vspace{-2mm}
\end{table}

\begin{table}[htbp]
\centering
\caption{Hardware Results}
\label{tab:hardware_results}
\small
\begin{tabular}{@{}lccc@{}}
\toprule
\textbf{Condition} & \textbf{Trials} & \begin{tabular}{c} \textbf{Boulder RL} \\ \textbf{(Ours)} \end{tabular} & \begin{tabular}{c} \textbf{Human} \\ \textbf{Operator} \end{tabular} \\
\midrule
Small/Soft  & 10 & 90\% & 90\% \\
Big/Soft    & 10 & 80\% & 90\% \\
Small/Hard  & 10 & 50\% & 80\% \\
Big/Hard    & 10 & 60\% & 70\% \\
\midrule
\textbf{Overall} & \textbf{40} & \textbf{70\%} & \textbf{83\%} \\
\bottomrule
\end{tabular}
\vspace{-2mm}
\end{table}

  The autonomous system achieved 70\% success rate compared to 83\% for the human operator (Table~\ref{tab:hardware_results}). Both performed
   best on small rocks in soft soil (90\% success), while small rocks in hard soil proved most challenging
  for the autonomous system (50\% vs 80\% human).

  \subsubsection{Comparative Failure Analysis}

  Human operators achieve their 83\% success rate using vision alone, interpreting soil deformation,
  bucket motion, and rock movement to infer resistance and adjust strategies accordingly. The autonomous
  system achieves 70\% success using sparse point clouds and proprioceptive feedback.

  Both systems exhibited similar failure modes in hard soil conditions: rocks sliding along the ground
  surface rather than entering the bucket when insufficient penetration depth prevented proper scooping. In
  soft soil, failures occurred when the bucket was overfilled, causing rocks to fall during lifting.

  The largest performance gap (30\%) occurred with small rocks in hard soil. Human operators maintained
  visual tracking through occlusions, compensating with spatial memory when rocks disappeared behind the
  bucket. In contrast, our vision system loses tracking during such occlusions, and small rocks are
  particularly susceptible to complete visual obstruction during the final approach. The cabin-mounted
  sensors compound this issue, as the arm itself blocks the camera's view during close manipulation.
  Additionally, the autonomous system struggles to recover when debris accumulates or soil height varies
  around the rock—conditions under which humans detect changes and intuitively reposition for cleaner
  approaches.

  \subsection{Soil-Adaptive Behavior}

\begin{figure}[htbp]
  \vspace{-2mm}
      \centering
      \includegraphics[width=\linewidth]{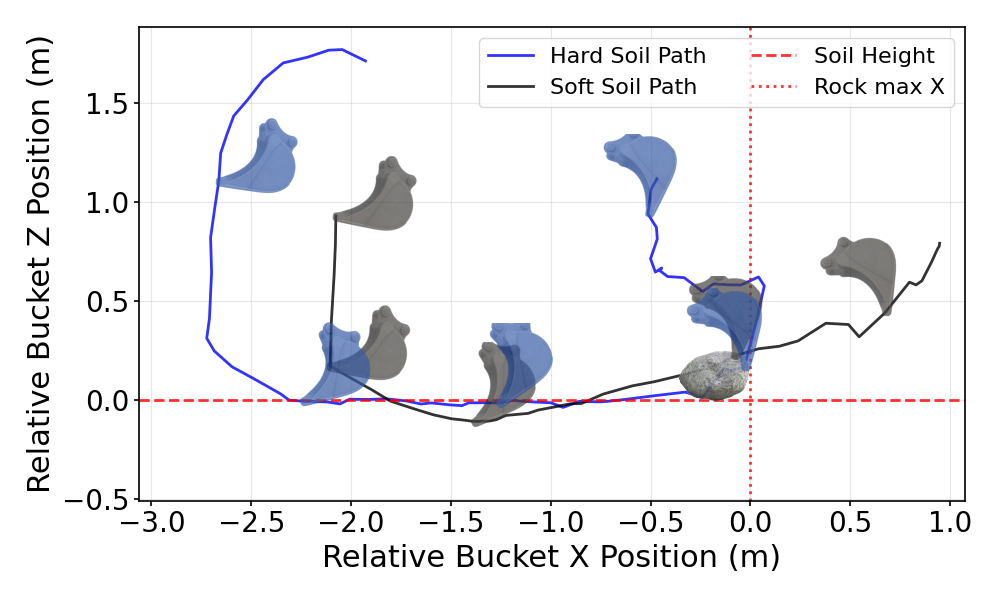}
      \caption{Shovel trajectories in hard vs soft soil conditions. The policy adapts its strategy based on
  resistance: extended surface dragging in hard soil versus direct penetration in soft soil. Trajectories
  aligned at soil level for comparison.}
      \label{fig:shovel_path_data}
  \vspace{-3mm}
  \end{figure}

  The policy exhibits different excavation strategies based on soil resistance, consistent with behaviors
  observed in prior soil-adaptive excavation work~\cite{egli2022soil}. Fig.~\ref{fig:shovel_path_data} shows
   representative trajectories from hardware deployment. In hard soil, the bucket follows an extended
  horizontal path (approximately 40\% longer), dragging along the surface to minimize penetration
  resistance. In soft soil, the bucket penetrates more directly, reducing the horizontal scooping distance.

  These strategies emerge from the interaction between proprioceptive feedback and the reward structure,
  without explicit soil property estimation. Dragging behavior is observed when torque readings are high
  during initial penetration attempts, while more direct approaches occur when resistance is low. Human
  operators develop similar strategies, but based almost entirely on visual cues, observing how soil deforms
   and how the bucket progresses to infer resistance levels.

  \subsection{Control Behavior Analysis}

\begin{figure}[htbp]
  \vspace{-3mm}
    \centering
    \begin{subfigure}[t]{0.9\columnwidth}
      \centering
      \includegraphics[width=\linewidth]{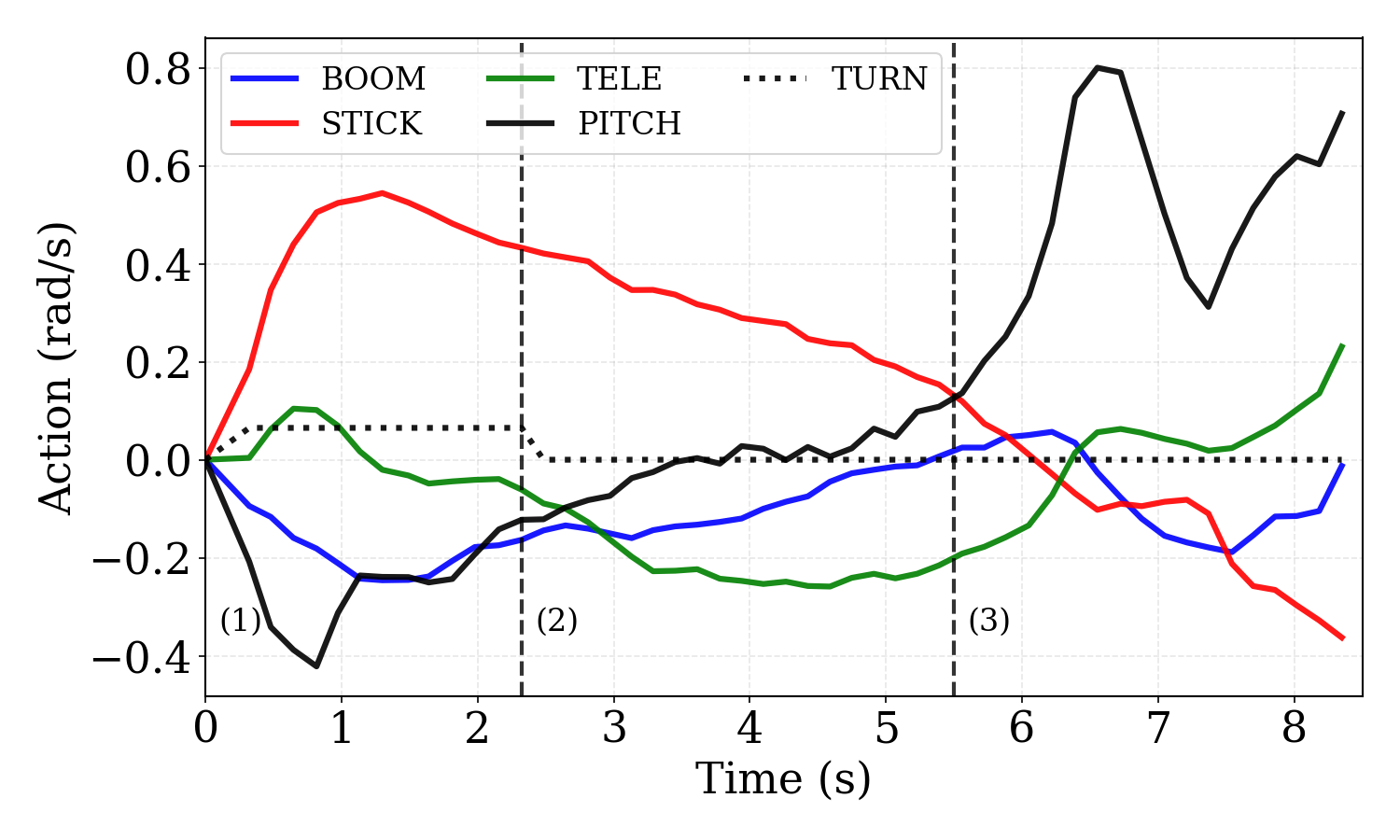}
      \caption{Joint velocity commands}
      \label{fig:jointactions}
    \end{subfigure}
    \begin{subfigure}[t]{0.9\columnwidth}
      \centering
      \includegraphics[width=\linewidth]{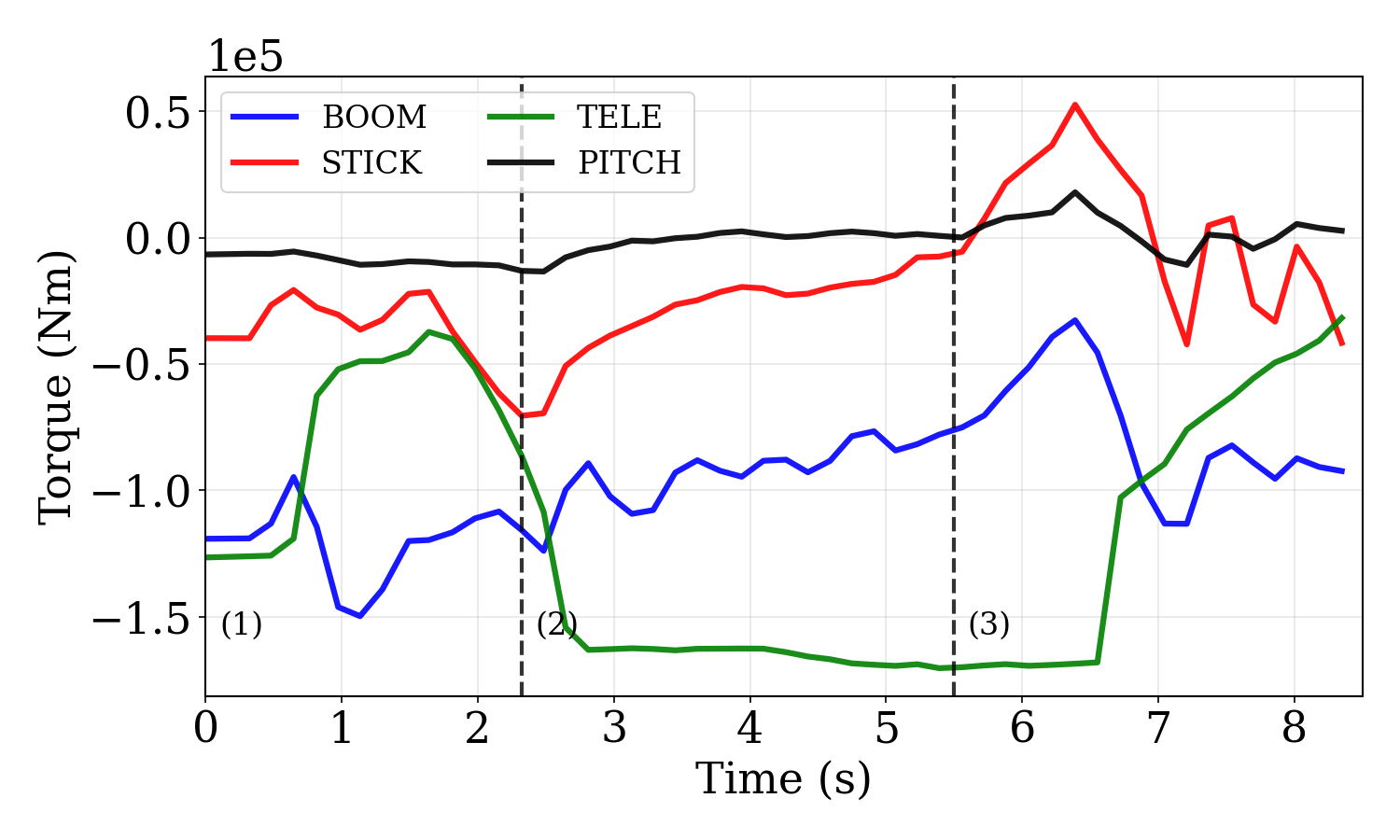}
      \caption{Measured joint torques (turn joint excluded)}
      \label{fig:jointtorques}
    \end{subfigure}
    \caption{Control signals during successful rock extraction. Vertical lines indicate phase transitions:
  alignment $\rightarrow$ scooping $\rightarrow$ lifting.}
    \label{fig:actionsandtorques}
  \vspace{-3mm}
  \end{figure}

  Fig.~\ref{fig:actionsandtorques} presents joint commands and torque measurements during successful
  execution. The control sequence shows three distinct emergent phases:

  \vspace{-1mm}
  \begin{enumerate}
  \setlength{\itemsep}{0pt}
  \item \textbf{Alignment}: Turn joint actively positions the rock in front of the bucket
  \item \textbf{Scooping}: Coordinated boom, stick, and bucket motion to capture the rock
  \item \textbf{Lifting}: Upward motion with bucket curl to secure the load
  \end{enumerate}
  \vspace{-1mm}

  Once aligned, turn commands consistently enter the dead zone (Eq.~\eqref{eq:deadzone}), effectively
  constraining the task to 2D manipulation. This mirrors human operation where operators typically align
  first, then execute scooping in a vertical plane. The torque profiles reveal how soil conditions influence
   execution: elevated stick and boom torques in hard soil (Fig.~\ref{fig:jointtorques}) correlate with the
  extended dragging trajectories.

  \subsection{Failure Modes and Real-World Challenges}

  Analysis of failed trials reveals three primary categories where simulation assumptions break down:

  \vspace{-1mm}
  \begin{itemize}
  \setlength{\itemsep}{0pt}
  \item \textbf{Perception and occlusion}: The fixed cabin-mounted sensors create blind spots during close
  manipulation. This manifests in two ways: (1) complete tracking loss when small rocks disappear behind the
   bucket, and (2) inability to detect partially buried rocks where only the visible portion is captured by
  LiDAR. These perception failures account for the majority of unsuccessful trials, particularly with small
  rocks.

  \item \textbf{Terrain and soil dynamics}: Real environments introduce unmodeled boulder-soil coupling that affects extraction dynamics, while accumulated soil progressively fills the bucket, reducing space for rock capture and causing spillage failures observed in soft soil conditions. 
  
  \item \textbf{Control constraints}: Imprecise turn control (due to coarse dynamics modeling) sometimes
  leads to suboptimal alignment with large rocks, occasionally causing lateral entrapment against the
  bucket.

  \end{itemize}
  \vspace{-1mm}

  These limitations highlight fundamental trade-offs in sim-to-real transfer: while higher-fidelity
  simulation, such as particle-based methods, could address some issues, the computational cost currently
  prohibits large-scale RL training.

\section{Conclusion}

We demonstrated autonomous boulder extraction on a 12-ton hydraulic excavator, achieving 70\% success rate compared to 83\% for human operators across rocks (0.4--0.7m) and varying soil conditions. The reinforcement learning policy, trained with rigid-body simulation and analytical soil models, adapts its strategy to soil resistance by dragging along hard surfaces and penetrating directly in soft conditions, using only standard digging buckets.

The system operates with 20 LiDAR points per rock and proprioceptive feedback, showing that boulder manipulation does not require dense geometric reconstruction. Adaptive behaviors emerge from the interaction between proprioceptive sensing and reward structure without explicit soil property estimation. Primary failures occur when small rocks become fully occluded during approach, causing tracking loss. The fixed cabin-mounted sensors create blind spots during close manipulation, and the analytical soil model cannot capture all real-world boulder-soil coupling effects.

This work shows that boulder manipulation, which previously required tool changes or manual operation, can be automated using standard excavator buckets through learning-based control. The approach demonstrates that practical construction automation often requires adaptive behavior rather than sophisticated hardware. Future work should address tracking through occlusions using temporal models and extend to multi-rock scenarios common in site clearing.
\bibliographystyle{IEEEtran}
\bibliography{references}
\end{document}